\documentclass[11pt]{article}
\usepackage[utf8]{inputenc}
\usepackage{acl2016}
\usepackage{times}
\usepackage{url}
\usepackage{latexsym}
\usepackage{tipa}
\usepackage{paralist}
\usepackage{multirow}
\usepackage{graphicx}
\usepackage{color}
\usepackage{tipa}

\aclfinalcopy 

\title{Siamese convolutional networks based on phonetic features for cognate identification}

\author{Taraka Rama \\
  Seminar f\"ur Sprachwissenschaft \\
  University of T\"ubingen \\
  {\tt taraka-rama.kasicheyanula@uni-tuebingen.de}}

\date{}

\begin{document}
\maketitle
\begin{abstract}
In this paper, we explore the use of convolutional networks 
(ConvNets) for the purpose of cognate identification. We compare our architecture with binary 
classifiers based on string similarity measures on different language families. Our experiments 
show that convolutional networks achieve competitive results across concepts and across language 
families at the task of cognate identification. 
\end{abstract}

\section{Introduction}
Cognates are words that are known to have descended from a common ancestral language. In historical 
linguistics, identification of cognates is an important step for positing relationships between 
languages. Historical linguists apply the comparative method \cite{trask1996historical} for 
positing relationships between languages.

In NLP, automatic identification of cognates is associated with the task of determining if two 
words are descended from a common ancestor or not. There are at least two ways to achieve automatic 
identification of cognates.

One way is to modify a well-known string alignment technique such as 
Longest Common Subsequence or Needleman-Wunsch algorithm \cite{needleman1970general} to weigh the 
alignments differentially \cite{kondrak2001identifying,list2012sca}. The weights are determined 
through the linguistic knowledge of the sound changes that occurred in the 
language family.

The second approach employs a machine learning perspective that is widely employed 
in NLP. The cognate identification is achieved by training a linear classifier or a sequence 
labeler on a set of labeled positive and negative examples; and then employ the trained classifier 
to classify new word pairs. The features for a classifier consist of word similarity measures 
based on number of shared bigrams, edit distance, and longest common subsequence 
\cite{hauer-kondrak:2011:IJCNLP-2011,inkpen2005automatic}.


The above procedures provide an estimate of the similarity between a pair of words and cannot 
directly be used to infer a phylogeny based on models of trait evolution. The 
pairwise judgments have to be converted into multiple 
cognate judgments so that the multiple judgments can be supplied to a automatic tree building 
program for inferring a phylogeny for the languages under study.

It has to be noted that the Indo-European dating studies 
\cite{bouckaert2012mapping,chang2015ancestry} employ human expert 
cognacy judgments for inferring phylogeny and dates of a very well-studied language family. Hence, 
there is a need for developing automated cognate 
identification methods that can be applied to under-studied languages of the world.

\section{Related work}
The earlier computational effort of \cite{jager2013phylogenetic,rama2013two} employs 
Pointwise Mutual Information (PMI) to compute transition matrices between sounds. 
Both \newcite{jager2013phylogenetic} and \newcite{rama2013two} employ undirectional sound 
correspondence based scorer to compute word similarity. The general approach is to align word pairs 
using vanilla edit distance and impose a cutoff to extract potential cognate pairs. The aligned 
sound symbols are then employed to compute the PMI scoring matrix that is used to realign the pairs. 
The PMI scoring matrix is recounted from the realigned pairs. This procedure is repeated until 
convergence. 

\newcite{jager2013phylogenetic} imposes an additional cutoff based on the PMI scoring matrix. 
Further, \newcite{jager2013phylogenetic} also employs the PMI scoring matrix to infer family trees 
for new language families and compares those trees with the \emph{expert} trees given in 
\emph{Glottolog} \cite{nordhoff2011glottolog}. \newcite{rama2013two} take a slightly different 
approach, in that, the authors compute a PMI matrix independently for each language family and 
evaluate its performance at the task of pair-wise cognate identification. In this work, we also 
compare the convolutional networks against PMI based binary classifier.

Previous works of cognate identification such as \cite{Bergsma:07,inkpen2005automatic} supply 
string similarity measures as features for training different classifiers such as decision trees, 
maximum-entropy, and SVMs for the purpose of determining if a given word pair is cognate or not. 

In another line of work, \newcite{list2012sca} employs a transition matrix derived from 
historical linguistic knowledge to align and score word pairs. This approach is algorithmically 
similar to that of \newcite{Kondrak:00} who employs articulation motivated weights to score a sound 
transition matrix. The weighted sound transition matrix is used to score a word pair.

The work of \newcite{list2012sca} known as Sound-Class Phonetic Alignment (SCA) approach reduces 
the phonemes to historical linguistic motivated sound classes such that transitions between some 
classes are less penalized than transitions between the rest of the classes. For example, the 
probability of velars transitioning to palatals is a well-attested sound change across the world. 
The SCA approach employs a weighted directed graph to model directionality and proportionality of 
sound changes between sound classes. For example, a direct change between velars and dentals is  
unattested and would get a zero weight. Both \newcite{Kondrak:00} and 
\newcite{list2012sca} set the weights and directions in the sound transition graph to suit the 
reality of sound change. 

All the above outlined approaches employ a scoring matrix that is derived automatically or manually; 
or, employ a SVM to train form similarity based features for the purpose of cognate identification.

\section{Convolutional networks}
This article is the first to apply convolutional networks (ConvNets) to phonemes by 
treating each phoneme as a vector of binary valued phonetic features. This approach has the 
advantage that it does not require explicit feature engineering, alignments,
and a sound transition matrix. The approach requires cognacy statements and phonetic descriptions of 
sounds used to transcribe the words. The cognacy statements can be obtained from etymological 
dictionaries and the quality of the phonemes can be obtained from \newcite{ladefoged1998sounds}.

\newcite{collobert2011natural} proposed ConvNets for NLP tasks in 2011 and were since applied for 
sentence classification 
\cite{kim:2014:EMNLP2014,DBLP:conf/naacl/Johnson015,KalchbrennerACL2014,NIPS2015_5782}, 
part-of-speech tagging \cite{santos2014learning}, and information retrieval \cite{shen2014latent}.

\newcite{kim:2014:EMNLP2014} applied convolutional networks to pre-trained word embeddings in a 
sentence for the task of sentence classification. \newcite{DBLP:conf/naacl/Johnson015} train their 
convolutional network from scratch by using a one-hot vector for each word. The authors show that 
their convolutional network performs better than a SVM classifier trained on bag-of-words features. 
\newcite{santos2014learning} use character embeddings to train their POS-tagger. The authors find 
that the POS-tagger performs better than the accuracies reported in \cite{manning2011part}.

In a recent work, \newcite{NIPS2015_5782} treat documents as a sequence of characters and transform 
each document into a sequence of one-hot character vectors. The authors designed and trained two 
9-layer convolutional networks for the purpose of sentiment classification. The authors report 
competitive or state-of-the art performance on a wide range of benchmark sentiment classification 
datasets.

\section{Character convolutional networks}
\newcite{chopra2005learning} extended the traditional ConvNets 
to classify if two images belong to the same person. These ConvNets are known as 
Siamese Networks (inspired from Siamese twins) and share weights for independent but identical 
layers of convolutional networks. Siamese networks and their variants have been employed for 
identifying if two images are from the same person or different persons 
\cite{Zagoruyko_2015_CVPR}; and for recognizing if two speech segments belong to the same word 
class \cite{DBLP:journals/corr/KamperWL15}.

\subsection{Word as image}
Historical linguists perform cognate identification based on regular correspondences which are 
described as changes in phonetic features of phonemes. For instance, Grimm's law $b^h \sim b$ is 
described as loss of aspiration; $p \sim f$ is described as change from plosives to fricatives; and 
devoicing $d \sim t$ in English \emph{ten} $\sim$ Latin \emph{decem}.

Learning criteria for cognacy through phonetic features 
from a set of training examples implies that there is no need for explicit alignment and 
design/learning of sound scoring matrices. In this article, we 
represent each phoneme as a binary-valued vector of phonetic features and then perform convolution 
on the two-dimensional matrix.

\subsection{Siamese network}
Intuitively, a network should learn a similarity function such that words that 
diverged due to accountable sound shifts are placed close to one another than two 
words that are not cognates. And, Siamese networks are suitable for this task since, they learn a 
similarity function that has a higher similarity between cognates as compared to non-cognates. The 
weight tying ensures that two cognate words sharing similar phonetic features in a local context 
tend to be get higher weights than words that are not cognate.

\subsection{Phoneme vectorization}
In this article, we work with the ASJP alphabet \cite{brown2013sound}. The ASJP alphabet is coarser 
than IPA but is designed with the aim 
to capture highly frequent sounds in the world's languages. The ASJP database has word 
lists for 60\% of the world's languages but only has cognate judgments for some selected families 
\cite{wichmann2013languages}.

\begin{table}[!ht]
\small
 \begin{tabular}{p{7cm}}
\hline
\hline
p = voiceless bilabial stop and fricative [IPA: p, \textipa{F}]\\
b = voiced bilabial stop and fricative [IPA: b, \textipa{B}] \\
m = bilabial nasal [IPA: m] \\
f = voiceless labiodental fricative [IPA: f] \\
v = voiced labiodental fricative [IPA: v] \\
8 = voiceless and voiced dental fricative [IPA: \textipa{T}, \textipa{D}] \\
4 = dental nasal [IPA: \textipa{\|[n}] \\
t = voiceless alveolar stop [IPA: t] \\
d = voiced alveolar stop [IPA: d] \\
s = voiceless alveolar fricative [IPA: s] \\
z = voiced alveolar fricative [IPA: z] \\
c = voiceless and voiced alveolar affricate [IPA: ts, dz] \\
n = voiceless and voiced alveolar nasal [IPA: n] \\
S = voiceless postalveolar fricative [IPA: \textipa{S}] \\
Z = voiced postalveolar fricative [IPA: \textipa{Z}] \\
C = voiceless palato-alveolar affricate [IPA: t\textipa{S}] \\
j = voiced palato-alveolar affricate [IPA: d\textipa{Z}] \\
T = voiceless and voiced palatal stop [IPA: c, \textipa{\textbardotlessj}] \\
5 = palatal nasal [IPA: \textipa{\textltailn}]\\
k = voiceless velar stop [IPA: k]\\
g = voiced velar stop [IPA: g]\\
x = voiceless and voiced velar fricative [IPA: x, \textipa{G}]\\
N = velar nasal [IPA: \textipa{N}]\\
q = voiceless uvular stop [IPA: q]\\
G = voiced uvular stop [IPA: \textipa{\;G}]\\
X = voiceless and voiced uvular fricative, voiceless and voiced pharyngeal fricative [IPA:
\textipa{X}, \textipa{K}, \textipa{\textcrh}, \textipa{Q}]\\
7 = voiceless glottal stop [IPA: \textipa{P}]\\
h = voiceless and voiced glottal fricative [IPA: h, \textipa{H}]\\
l = voiced alveolar lateral approximate [IPA: l]\\
L = all other laterals [IPA: L, \textipa{L}]\\
w = voiced bilabial-velar approximant [IPA: w]\\
y = palatal approximant [IPA: j]\\
r = voiced apico-alveolar trill and all varieties of ``r-sounds'' [IPA: r, R, etc.]\\
! = all varieties of ``click-sounds'' [IPA: !, \textipa{\!o}, \textipa{||}, 
\textipa{\textdoublebarpipe}]\\
\hline
 \end{tabular}
\caption{ASJP consonants. ASJP has 6 vowels which we collapsed to a single vowel 
\emph{V}.}
\label{tab:asjpconso}
\end{table}

We composed a binary vector for each phoneme based on the description given in table 
\ref{tab:asjpconso}. In total, there are 16 binary valued features. We also reduced all vowels to a 
single vowel that has a value of $1$ for voicing feature and $0$ for the rest of the features. The 
main motivation 
for such decision is that vowels are diachronically unstable than consonants 
\cite{kessler2007word}.

A word such as ``fat'' would be 
represented as $3 \times 16$ matrix where each column provides a binary value for the phonetic 
feature (cf. table \ref{tab:binary}).

\begin{table}[!ht]
\small
 \begin{tabular}{p{1.5cm}||p{4cm}}
\hline
\hline
p& 0 1 0 0 0 0 0 0 1 1 0 0 0 0 0 0\\
b& 1 1 0 0 0 0 0 0 1 1 0 0 0 0 0 0\\
f &0 1 1 0 0 0 0 0 0 1 0 0 0 0 0 0\\
v &1 1 1 0 0 0 0 0 0 1 0 0 0 0 0 0\\
m &1 1 0 0 0 0 0 0 0 0 0 1 0 0 0 0\\
8 &1 0 1 0 0 0 0 0 0 1 0 0 0 0 0 0\\
4 &1 0 1 0 0 0 0 0 0 0 0 1 0 0 0 0\\
t &0 0 0 1 0 0 0 0 1 0 0 0 0 0 0 0\\
d &1 0 0 1 0 0 0 0 1 0 0 0 0 0 0 0\\
s &0 0 0 1 0 0 0 0 0 1 0 0 0 0 0 0\\
z &1 0 0 1 0 0 0 0 0 1 0 0 0 0 0 0\\
c &1 0 0 1 0 0 0 0 0 0 1 0 0 0 0 0\\
n &1 0 0 1 0 0 0 0 0 0 0 1 0 0 0 0\\
S &0 0 0 0 1 0 0 0 0 1 0 0 0 0 0 0\\
Z &1 0 0 0 1 0 0 0 0 1 0 0 0 0 0 0\\
C &0 0 0 0 1 0 0 0 0 0 1 0 0 0 0 0\\
j &1 0 0 0 1 0 0 0 0 0 1 0 0 0 0 0\\
T &1 0 0 0 1 0 0 0 1 0 0 0 0 0 0 0\\
5 &0 0 0 0 1 0 0 0 0 0 0 1 0 0 0 0\\
k &0 0 0 0 0 1 0 0 1 0 0 0 0 0 0 0\\
g &1 0 0 0 0 1 0 0 1 0 0 0 0 0 0 0\\
x &1 0 0 0 0 1 0 0 0 1 0 0 0 0 0 0\\
N &1 0 0 0 0 1 0 0 0 0 0 1 0 0 0 0\\
q &0 0 0 0 0 0 1 0 1 0 0 0 0 0 0 0\\
G &1 0 0 0 0 0 1 0 1 0 0 0 0 0 0 0\\
X &1 0 0 0 0 0 1 0 0 1 0 0 0 0 0 0\\
7 &0 0 0 0 0 0 0 1 1 0 0 0 0 0 0 0\\
h &1 0 0 0 0 0 0 1 0 1 0 0 0 0 0 0\\
l &1 0 0 0 0 0 0 0 0 0 0 0 0 1 1 0\\
L &1 0 0 0 0 0 0 0 0 0 0 0 0 0 1 0\\
w &1 1 0 0 0 1 0 0 0 0 0 0 0 1 0 0\\
y &1 0 0 0 1 0 0 0 0 0 0 0 0 1 0 0\\
r &1 0 0 0 0 0 0 0 0 0 0 0 0 0 0 1\\
! &1 0 0 0 0 0 0 0 0 0 0 0 1 0 0 0\\
V &1 0 0 0 0 0 0 0 0 0 0 0 0 0 0 0\\\hline
 \end{tabular}
\caption{Binarized ASJP alphabet used in our experiments. Each column corresponds to 
the following features: Voiced, Labial, Dental, Alveolar, Palatal/Post-alveolar, Velar, Uvular, 
Glottal, Stop, Fricative, Affricate, Nasal, Click, Approximant, Lateral, and Rhotic.}
\label{tab:binary}
\end{table}

\subsection{ConvNet Models}
In this subsection, we describe the ConvNet models used in our experiments.

\begin{figure}
\centering
 \includegraphics[scale=0.5]{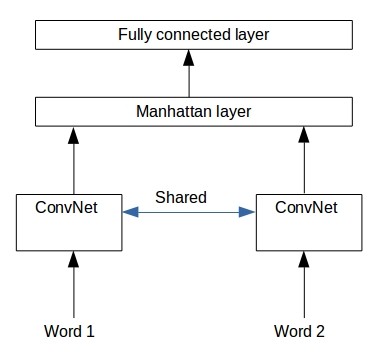}
\caption{Siamese network with fully connected layer. The weighted are shared between the two 
convolutional networks.}
\label{fig:siam}
\end{figure}

\textbf{Siamese ConvNet} Siamese 
networks takes a pair of inputs and minimizes the distance between the 
output representations. Each branch of the Siamese network is composed of a convolutional 
network. The Euclidean distance $D$ between the representations of each branch is then used to 
train a contrastive-loss function $y D + (1-y) max\{0, m-D\}$ where $m$ is the margin and $y$ is 
the true label. We only describe the architecture since this forms the basis for the rest of our 
experiments with Siamese architectures.\footnote{The results were slightly better than a majority 
class classifier and were not reported in the article}

\textbf{Manhattan Siamese ConvNet} The second ConvNet is also a Siamese network where the 
Euclidean distance is replaced by a element-wise absolute difference layer followed by a fully 
connected layer (cf. figure \ref{fig:siam}). To the best of our knowledge, 
only \newcite{Zagoruyko_2015_CVPR} added two fully connected layers to 
the concatenated outputs of the Siamese network and trained a system that predicts if two image 
patches belong 
to the same image or different images. We refer this architecture as a Manhattan Siamese ConvNet 
due to the difference layer's similarity to Manhattan distance.

\textbf{2-channel Convnet} Until now, each word is treated as a separate 
input. \newcite{Zagoruyko_2015_CVPR} introduced a 2-channel architecture which treats a pair of 
image patches as a 2-channel image. This can also be applied to words. The 2-channel 
ConvNet has two convolutional layers, a maxpooling layer, and a fully 
connected layer with 8 units.

The number of feature maps in each convolutional layer is fixed at $10$ with a kernel size of 
$2\times 3$. The max-pooling layer halves the output of the previous convolutional layer. We 
also inserted a dropout layer with $0.5$ probability \cite{srivastava2014dropout} after a 
fully-connected layer to 
avoid over-fitting. The convolutional layers were trained with ReLU non-linearity.

We zero-padded each word to obtain a length of $10$ for all the words to apply the filter equally 
about a word.
We used adadelta optimizer \cite{zeiler2012adadelta} with learning 
rate of $1.0$, $\rho = 0.95$, and $\epsilon = 10^{-6}$. We fixed the mini-batch size to 
$128$ in all our experiments. We experimented with different batch sizes ($[32, 64, 128, 256]$)  and 
did not observe any significant deviation in the validation loss. Both, Manhattan and 2-stream 
ConvNets were trained using the log-loss function. Both our architectures are relatively 
shallow (3) as compared to the text classification architecture of \newcite{NIPS2015_5782}. We 
trained all our networks using Keras \cite{chollet2015keras} and Theano \cite{bergstra2010theano}.



\section{Comparison methods}
We compare the ConvNet architectures with SVM classifiers trained with different string 
similarities as features.

\textbf{Other sound classes/alphabets} Apart from ASJP alphabet, there are two other alphabets 
that have been designed by historical linguists for the purpose of modeling sound change. As 
mentioned before, the main idea behind the design of sound classes is to discourage transitions 
between particular classes of sounds but allow transitions within a sound class. 
\newcite{dolgopolsky1986probabilistic} proposed a ten sound class system based on the empirical data 
of $140$ languages. SCA alphabet \cite{list2012sca} has a size of $25$ and attempts to 
address some issues with the ASJP alphabet (lack of tones) and also extend Dolgopolsky's sound 
classes based on evidence from more number of languages.

\textbf{Orthographic measures as features} We converted all the datasets into all the three sound 
classes and computed the following string similarity scores:
\begin{compactitem}
 \item Edit distance.
\item Common number of bigrams.
\item Length of the longest common subsequence.
\item Length of longest common prefix.
\item Common number of trigrams.
\item Global alignment based on Needlman-Wunch algorithm \cite{needleman1970general}.
\item Local alignment score based on Smith-Waterman algorithm \cite{smith1981identification}.
\item Semi-global alignment score is a compromise between global and local alignments  
\cite{durbin1998biological}.\footnote{The global, local, and alignment scores were computed using 
LingPy library \cite{list-moran:2013:SystemDemo}.}
\item Common number of skipped bigrams (XDICE).
\item A positional extension of XDICE known as XXDICE \cite{brew1996word}.
\end{compactitem}

\textbf{Pair-wise Mutual Information (PMI)} We also computed a PMI score for a pair of ASJP 
transcribed words using the PMI scoring matrix developed by \newcite{jager2013phylogenetic}. This 
system is referred to as PMI system.

We included length of each word and the absolute difference in length between the words as 
features for both the Orthographic and PMI systems. The sound class orthographic scores 
system attempts to combine the previous cognate identification systems developed by 
\cite{inkpen2005automatic,hauer-kondrak:2011:IJCNLP-2011} and the insights from applying 
string similarities to sound classes for language comparison \cite{kessler2007word}.

\section{Datasets}
In this section, we describe the datasets used in our experiments.

\textbf{IELex database} The Indo-European Lexical database is created by
\newcite{dyen1992indoeuropean} and curated by Michael Dunn.\footnote{\url{ielex.mpi.nl}} The 
transcription in IELex database is not uniformly IPA and retains many forms transcribed in 
the Romanized IPA format of \newcite{dyen1992indoeuropean}. We cleaned the IELex database of any 
non-IPA-like transcriptions and converted part of the database into ASJP 
format.

\textbf{Austronesian vocabulary database} The Austronesian Vocabulary 
Database \cite{greenhill2009austronesian} has word lists for 210 Swadesh concepts and 378 
languages.\footnote{\url{http://language.psy.auckland.ac.nz/austronesian/}} The database 
does not have transcriptions in a uniform IPA format. We 
removed all symbols that do not appear in the standard IPA and converted the lexical items to ASJP 
format.\footnote{For computational reasons, we work with a subset of 100 languages.}

\begin{table}[!ht]
\footnotesize
\centering
\begin{tabular}{p{1.6cm}p{1.cm}p{1.cm}p{1.cm}p{1.2cm}}
\hline
Family & Concepts & Languages & Training & Testing\\\hline
Austronesian & $210$ & $100$ & $334807$ & $140697$ \\
Mayan & $100$ & $30$ & $28222$ & $12344$\\
Indo-European & $206$ & $50$ & $117740$ & $49205$\\ 
Mixed dataset & -- & -- & $176889$ & -- \\
\hline
\end{tabular}
\caption{The number of languages, concepts, training, and test examples in our datasets. We do not 
test on the mixed database and only use it for training purpose.}
\label{tab:data}
\end{table}

\textbf{Short word lists with cognacy judgments} \newcite{wichmann2013languages} and 
\newcite{List2014d} compiled cognacy wordlists for subsets of families from 
various scholarly sources such as comparative handbooks and historical linguistics' articles. 
The details of this compilation is given below. For each dataset, we give the number of 
languages/the number of concepts in parantheses. This dataset is henceforth referred to as ``Mixed 
dataset''.

\begin{itemize}
 \item \newcite{wichmann2013languages}: Afrasian (21/40), Kadai (12/40), Kamasau (8/36), 
Lolo-Burmese (15/40), Mayan (30/100), Miao-Yao (6/36), Mixe-Zoque (10/100), Mon-Khmer (16/100).
\item \newcite{List2014d}: Bai dialects (9/110), Chinese dialects (18/180), Huon (14/84), Japanese 
(10/200), ObUgrian (21/110; Hungarian excluded from Ugric sub-family), Tujia (5/107; Sino-Tibetan).
\end{itemize}

We performed two experiments with these datasets. In the first experiment, we randomly selected 
70\% of concepts from IELex, ABVD, and Mayan datasets for training and the rest of the 30\% 
concepts for testing.  The motivation behind this experiment is to test if ConvNets can learn 
phonetic feature patterns across concepts. In the second experiment, we trained on the Mixed 
dataset but tested on the Indo-European and Austronesian datasets. The motivation behind this 
experiment is to test if ConvNets can learn general patterns of sound change across language 
families. The number of training and testing examples in each dataset is given in table 
\ref{tab:data}.

\begin{table*}[!t]
\centering
\begin{tabular}{lccccc}\hline
Language family & Orthographic & PMI & Manhattan ConvNet & 2-Channel ConvNet
\\\hline
Austronesian & $77.92\%$ & $78\%$ & $79.04\%$ & $76.1\%$ \\
Indo-European & $80\%$ & $78.58\%$ &  $83.43\%$ & $81.7\%$ \\
Mayan & $83.66\%$ & $85.25\%$ &  $87.1\%$ & $82.1\%$ \\\hline
\hline
\multirow{3}{*}{Austronesian} & $0.833$ & $0.836$ & $0.861$ & $0.830$\\
& $0.675$ & $0.665$ & $0.576$ & $0.595$\\
& $0.783$ & $0.782$ & $0.776$ & $0.760$ \\\hline
\multirow{3}{*}{Indo-European} & $0.863$ & $0.854$ & $0.894$ & $0.883$ \\
& $0.628$ & $0.598$ & $0.618$ & $0.585$\\
& $0.808$ & $0.794$ & $0.830$ & $0.813$\\\hline
\multirow{3}{*}{Mayan} & $0.866$ & $0.885$ &$0.888$ & $0.865$\\
&$0.791$ & $0.795$&$0.756$ & $0.734$\\
& $0.84$& $0.853$ & $0.842$ & $0.819$\\\hline
\hline 
Austronesian & $0.749$ & $0.74$  & $0.683$ & $0.643$ \\
Indo-European & $0.729$ & $0.678$  & $0.681$ & $0.64$ \\
Mayan & $0.88$ & $0.892$ &  $0.871$ & $0.805$ \\\hline
\hline
\end{tabular}
\caption{Each system is trained on cognate and non-cognate pairs on 145 concepts in Indo-European 
and Austronesian families; and tested on the rest of the concepts. For Mayan family, the 
number of training concepts is 70 and the number of concepts in testing data is 30. For each 
family, numbers correspond to the following metrics: Accuracies, F-scores 
(negative, positive, combined), Average precision score.}
\label{tab:scores}
\end{table*}

\section{Results}
In this section, we report the results of our cross-concepts and cross-family experiments.

\textbf{SVM training and evaluation metrics} We used a linear kernel and optimized the SVM 
hyperparameter ($C$) through ten-fold cross-validation and grid search on the training data. We 
report accuracies, class-wise F-scores (positive and negative), combined F-score, and average 
precision score for each system on concepts dataset in table \ref{tab:scores}. The average precision 
score corresponds to the area under the precision-recall curve and is an indicator of the robustness 
of the model to thresholds.

\subsection{Cross-Concept experiments}
\textbf{Effect of size and width of fully connected layers} We observed that both the depth and 
width of the fully connected layers do not affect the performance of the ConvNet models. 
We used a fully connected network of size 8 in all our experiments. We increased the number of 
neurons from $8$ to $64$ in multiples of two and observed that increasing the number of neurons 
hurts the performance of the system.

\textbf{Effect of filter size} \newcite{zhang2015sensitivity} observed that the size of the filter 
patch can affect the performance of the system. We experimented with different filter sizes of 
dimensions $m \times k$ where, $m \in [1,2]$ and $k \in [1,3]$. We did not find any change in the 
performance in concepts experiments. We report the results for $m=2, k=3$ filter size for 
cross-concept experiments.

\subsection{Cross-Family experiments}
\textbf{Effect of filter size} Unlike the previous experiment, the filter size has a effect on the 
performance of the ConvNet system. We observed that the best results were obtained with a filter 
size of $1\times 3$. 

We did not include the results of the 2-channel ConvNet because of its worse performance at the 
task of cross-family cognate identification. The results of our experiments are given in table 
\ref{tab:lngscores}.

\begin{table}[ht!]
\small
\centering
\begin{tabular}{lccp{1.7cm}}\hline
Dataset & Orthographic & PMI & Manhattan ConvNet\\\hline
Austronesian & $0.766$ & $0.78$ & $0.746$ \\
Indo-European & $0.815$ & $0.804$ &  $0.804$ \\\hline
\hline
\multirow{3}{*}{Austronesian} & $0.821$ & $0.837$ & $0.820$\\
                             & $0.661$ & $0.656$ & $0.570$\\
                             & $0.759$ & $0.768$ & $0.728$ \\\hline
\multirow{3}{*}{Indo-European} & $0.876$ & $0.873$ & $0.871$ \\
                               & $0.631$ & $0.569$ & $0.590$ \\
                               & $0.806$ & $0.786$ &  $0.791$ \\\hline\hline
Austronesian & $0.771$ & $0.795$ & $0.707$ \\
Indo-European & $0.731$ & $0.692$ &  $0.691$ \\\hline
\end{tabular}
\caption{Testing accuracies, class-wise and combined F-scores, average precision score of 
each system on Indo-European and Austronesian families.}
\label{tab:lngscores}
\end{table}

\section{Discussion}
The Manhattan ConvNet competes with PMI and orthographic models at cross-concept cognate 
identification task. The Manhattan ConvNet performs better than PMI and orthographic models in 
terms of overall accuracy in all the three language families. In terms of averaged F-scores, 
Manhattan ConvNet performs slightly better than orthographic model and only performs worse than the 
other models at Austronesian language family.

The Manhattan ConvNet shows mixed performance at the task of cross-family cognate identification. 
The Manhattan ConvNet does not turn up as the best system across all the evaluation metrics in a 
single language family. The ConvNet performs better than PMI but is not as good as Orthographic 
measures at Indo-European language family. In terms of accuracies, the ConvNet comes closer to PMI 
than the orthographic system.

These experiments suggest that ConvNets can compete with a classifier trained on 
different orthographic measures and different sound classes. ConvNets can also compete with a data 
driven method like PMI which was trained in an EM-like fashion on millions of word pairs. ConvNets 
can certainly perform better than a classifier trained on word similarity scores at cross-concept 
experiments.

The Orthographic system and PMI system show similar performance at the Austronesian cross-concept 
task. However, ConvNets do not perform as well as orthographic and PMI systems. The reason for this 
could be due to the differential transcriptions in the database.

\section{Conclusion}
In this article, we explored the use of phonetic feature convolutional networks for 
the task of pairwise cognate identification. Our experiments with convolutional networks 
show that phonetic features can be directly used for classifying if two words are related or not. 

In the future, we intend to work directly with speech recordings and include language relatedness 
information into ConvNets to improve the performance. We are currently working towards building a 
larger database of word lists in IPA transcription.

\section*{Acknowledgments}
I thank Aparna Subhakari, Vijayaditya Peddinti, Johann-Mattis List, Johannes Dellert, Armin Buch, 
Çağrı Çöltekin, Gerhard Jäger, and Daniël de Kok for all the useful comments. The data for the 
experiments was processed by Johann-Mattis List and Pavel Sofroniev.

\bibliography{myreflnks}
\bibliographystyle{acl2016}



\end{document}